\documentclass[12pt]{article} 

\usepackage[utf8]{inputenc} 

\usepackage{physics}
\usepackage{graphicx}
\graphicspath{ {./images/} }
\usepackage{geometry} 
\usepackage{graphicx} 

\usepackage{booktabs} 
\usepackage{array} 
\usepackage{paralist} 
\usepackage{verbatim} 
\usepackage{subfig} 
\usepackage{fancyhdr} 
\usepackage[detect-all]{siunitx}
\usepackage{pgfplots}
\usepackage{sectsty}
\allsectionsfont{\sffamily\mdseries\upshape} 

\usepackage[nottoc,notlof,notlot]{tocbibind} 
\usepackage[titles,subfigure]{tocloft} 

\title{An Amalgamation of Classical and Quantum Machine Learning For the Classification of Adenocarcinoma and Squamous Cell Carcinoma Patients}
\author{Siddhant Jain$^{1,2}$, Jalal Ziauddin$^{2}$, Paul Leonchyk$^{2}$, Joseph Geraci$^{2,3}$}

\date{%
    $^1$University of Toronto\\%
    $^2$NetraMark Corp\\%
    $^3$Queen's University, Molecular Medicine\\[2ex]%
    \today
}

\begin{document}
\maketitle

\section* {Abstract}

The ability to accurately classify disease subtypes is of vital importance, especially in oncology where this capability could have a life saving impact. Here we report a classification between two subtypes of non-small cell lung cancer, namely Adenocarcinoma vs Squamous cell carcinoma.  The data consists of approximately 20,000 gene expression values for each of 104 patients. The data was curated from \cite{gse} \cite{gsee}.  We used an amalgamation of classical and and quantum machine learning models to successfully classify these patients. We utilized feature selection methods based on univariate statistics in addition to XGBoost \cite{boost}.  A novel and proprietary data representation method developed by one of the authors called QCrush was also used as it was designed to incorporate a maximal amount of information under the size constraints of the D-Wave quantum annealing computer. The machine learning was performed by a Quantum Boltzmann Machine. This paper will report our results, the various classical methods, and the quantum machine learning approach we utilized. 

\section {Introduction}
The primary purpose of this paper is to elaborate on the process we used to classify two specific lung cancers, namely Adenocarcinoma and Squamous Cell Carcinoma, from a 20,000 gene expression set using a combination of classical and quantum machine learning models. In the process of achieving consistent classification we utilized a process of iterative hyper-tuning. This paper will focus largely on the algorithms and procedures used to complete computations with the Quantum Processing Unit (QPU) provided by D-Wave to harness its unique ability to enhance specific operations used for our classification of lung cancers.

In the spirit of the work in \cite{neartermqc} we would like to disclose that there were no attempts at demonstrating some type of quantum supremacy here. It is believed that classical machine learning will dominate the landscape of data science for many years while serious quantum computational hardware challenges are met. However, it is believed that near term devices will be able to provide impressive advantages for certain tasks and that hybrid classical and quantum methods will be a way forward. With that in mind, we utilize the ability of quantum annealers to generate complex probability distributions to aid a machine learning protocol in the arena of medicine. 

\section{Precision Medicine and Machine Learning}
The idea that complex disorders like Alzheimer's disease, and many cancers, are in fact syndromes consisting of several sub-diseases is not new. However, our ability to better stratify patient populations for these disorders arrived with machine learning. Oncology led the way with what we refer to as biomarkers \cite{bio}, which are sets of variables that one can measure from a patient, like gene expression or cholesterol level, that can be used to make a prediction or determination about a person's health. For example, will a patient respond to some specific treatment, or is a patient at risk for having at heart attack in the near future? Terms like personalized and precision medicine are usually used interchangeably, but a preference in general has emerged for precision medicine. This is simply because the term personalized seems to imply a treatment designed for one person specifically, which may occur in the future, whereas the term precision is based on the idea that we could understand a true subtype that a patient lies within and for whom a particular treatment may be optimal. Machine learning is making this a reality and models already exist that influence cancer and other treatments \cite{precisionAI}. 

In addition to the very important work which focuses on making treatments for patients more precise, precision medicine through machine learning is also influencing drug design. The reason for wanting to do this is simple: if one says that they are going to design a drug for Alzheimer's, what do they really mean? What mechanism of action is this drug going to have? Machine learning is beginning to reveal that Alzheimer's and other disorders have multiple distinct manifestations that may likely require different interventions. Thus, work is commencing to precisely define these subtypes and what molecular machinery drives them, and this knowledge is going to influence the next generation of drug design. As quantum computers mature, quantum machine learning will play a role at both ends of this process. On the one hand, quantum computers provide a natural computational environment for the exploration of molecules, being that molecules are quantum mechanical structures. On the other hand, their ability to utilize quantum parallelism may be utilized to understand patient subpopulations. In the meanwhile, the machine learning protocols available for systems like the D-Wave 2000Q is allowing researchers to create patient response protocols that may one day influence clinical decisions. We present a unique effort in this paper by utilizing the D-Wave 2000Q computer to create a cancer biomarker, but there has already been some effort in the medical space including the work in \cite{lidar} which attempts to explore the potential impact that quantum machine learning may have on a computational biology problem. In addition to this, there have been various efforts in utilizing quantum annealers to model drugs and their interactions with proteins. One may refer to \cite{quantumcompanies} to learn about the many, albeit potentially premature yet exciting, commercial efforts commencing in this space.

\section {Quantum Computation and Quantum Processing Unit}
At the heart of a quantum computer is the Quantum Processing Unit (QPU) which provides a way to encode information in what is referred to as quantum bits (qubits) \cite{dwave}. A quantum bit, likened to a classical bit contains information that the machine uses to conduct operations. Unlike a classical bit the quantum bit does not default to a 0 or 1, rather it occupies a state that is in, what is referred to as, a superposition\cite{qbook} of 0 and 1 with an underlying probability\cite{qubit}. Once the qubit is measured by an observation, the qubit finds itself in a definite state of either 0 or 1, like an ordinary classical computer. The fact that the qubit, or more accurately, that the set of qubits within the computer, is in a superposition allows the machine to use quantum mechanics to evolve its state. This property, in addition to entanglement\cite{qbook}, produces a situation where the amount of information that can be simultaneously represented in one state of the quantum computer is exponentially greater than what can represented in a classical machine. In other words, if you have two qubits, a quantum computer can be in all 4 states $\ket{00}$, $\ket{01}$, $\ket{10}$, and $\ket{11}$ simultaneously while a classical computer can only be in one of these four states in any moment. This does mean that a quantum computing programmer has access to quantum mechanical properties and can evolve a state in a way that classical computers cannot, and thus perform some novel computations through quantum algorithms. This does not mean that quantum computers are in general exponentially faster, though they can perform some computations more efficiently. To be mathematically rigorous, if BQP are the class of problems that quantum computers can make a decision about in polynomial time, the the actual relationship between BQP and NP is not known\cite{qbook}.

Ultimately, the advent of quantum computers like the D-Wave computer represents the culmination of years of effort and the achievement is remarkable. Nevertheless, these modern quantum computers are still not superior in any practical way, however our ability to model lung cancer with such a machine indicates that as these machines scale, we will be able to access them to understand patient populations in a way not possible with classical computers. We will now review the D-Wave machine briefly.

\begin{figure}[h]
    \centering
\includegraphics[scale=0.43]{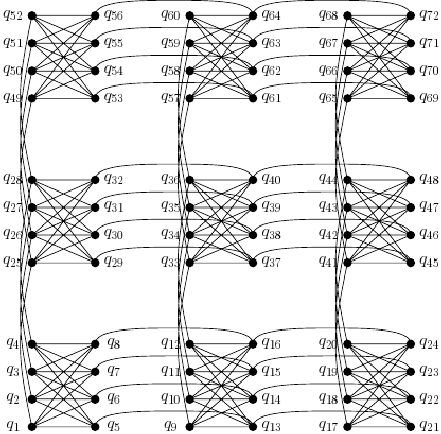}.
\caption{D-Wave Quantum Processing Unit Arranged by Unit Cells}
\label{chimera}
\end{figure}

The architecture of the Quantum Processing Unit used by D-Wave is a chimera graph as shown in figure \ref{chimera}.The paradigm of quantum computation that the D-Wave computer uses is known as annealing Quantum Computation or Quantum Annealing (QA) \cite{anneal}. Essentially, one first needs to understand that a problem can be encoded in an energy functional called a Hamiltonian. Next, note that in the D-Wave system one can start with a simple Hamiltonian in its ground state and then allow the technology to change its internal configuration in such a way so that it will end up in the ground state of the Hamiltonian that encodes the problem of interest. This has to be done in a very precise and time sensitive way, but is clearly possible\cite{Hamiltonian}. This allows us to solve a certain Hamiltonian required to learn about our cancer patients through Quantum Boltzmann Machines. See \cite{lidar} for more information on the QA paradigm. 

\section {Boltzmann Machine}
\begin{figure}[h]
    \centering
\includegraphics[scale=0.18]{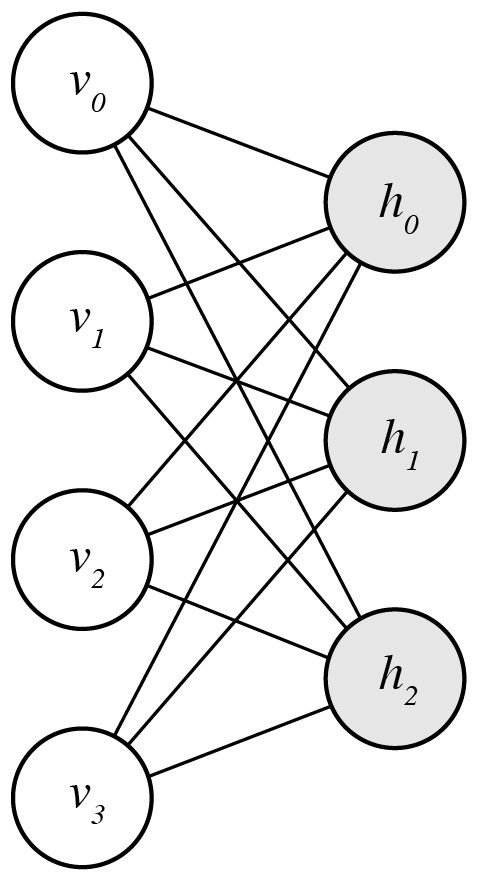}.
\caption{A Simple Restricted Boltzmann Machine Consisting of 4 Visible and 3 Hidden Nodes. Note the absence of intralayer edges.}
\label{rbm}
\end{figure}

A Boltzmann Machine is a generative machine learning model that is trained to encode the underlying distribution of a given dataset \cite{distr}. A Restricted Boltzmann Machine (RBM), a subset of Boltzmann Machines, is a bipartite graph that is segmented into visible and hidden neurons. The neurons of the visible layer are strongly connected to those in the hidden layer, but there exist no adjacent edges between neurons belonging to the same layer. We chose to use a RBM in our experiment because its training is less computationally expensive than a standard Boltzmann Machine \cite{orbm}. See figure \ref{rbm}.

Once trained, RBMs produce reconstructions of the provided data \cite{orbm}. One method of training, optimizing the weights of, an RBM is Contrastive Divergence, which is achieved by Gibbs sampling and gradient descent\cite{CD}. It is important to understand the workings of a Restricted Boltzmann Machine as it provides the foundation for understanding the training of Boltzmann machines on the D-Wave quantum annealer via the quantum sampler. Keep in mind that the idea to utilize this paradigm of machine learning was natural as the fundamental architecture of the Quantum Processing Unit closely resembles the graphical structure of an RBM.\cite{chimera} 

\subsection{Restricted Boltzmann Machine}
An RBM contains three parameters that encode the entire model: 
\begin{enumerate}[i)]
  \item the bias vector for the visible layer (length n)
  \item the bias vector for the hidden layer (length m)
  \item the weights matrix that represents the edge weight for each connection between the visible and hidden layer (m by n)
\end{enumerate} Once these three parameters are trained the RBM should be able to create reliable reconstructions of the data that it is provided. The significant hyper-parameters of a RBM are the learning rate and the number of hidden neurons. The learning rate simply determines the magnitude of change made to the parameters, and the number of hidden neurons determines the degree of information compression that occurs. By tuning the hyper-parameters you should be able to create a model that reconstructs the data provided with a high degree of accuracy.

The Boltzmann equation determines the energy of a system. In the context of an RBM this means the equation is as follows:

\begin{equation}
E(v,h) = -\sum_i a_i v_i - \sum_j b_j h_j -\sum_i \sum_j v_i w_{i,j} h_j
\end{equation}

The parameters that we seek to tune with training are the weights matrix, and the two bias vectors a and b. In matrix form:

\begin{equation}
E(v,h) = -a^{\mathrm{T}} v - b^{\mathrm{T}} h -v^{\mathrm{T}} W h
\end{equation}

To put this in perspective consider that energy based methods are designed to minimize an energy functional, as given by $E(v,h)$ above, for example. The proposed system is essentially trying to capture relationships between variables. Thus, if we train this system to minimize $E(v,h)$ then lower energy configurations will be favored by the system. This is because a standard way of computing the joint probability between $v$ and $h$ is given by 

\begin{equation}
P(v,h) = \frac{exp(-E(v,h))}{\sum_{v,h} exp(-E(v,h))}
\end{equation}

Thus, because we minimize $E(v,h)$, we end up maximizing 

\begin{equation}
P(v) = \frac{\sum_h exp(-E(v,h))}{\sum_{v,h} exp(-E(v,h))}
\end{equation}

Once this maximization occurs, we will have a system that will reflect the distribution of the training set.  We will not go into the technicalities of how this is accomplished here, however there are standard references available \cite{Hinton2012}. The essential idea for the reader to understand is that this system generalizes well on the D-Wave machine due to the architecture of the system. More specifically, the connectivity of the QPU given in figure \ref{chimera}, can be utilized to define a system that is the quantum analogy of the Restricted Boltzmann Machines just described \cite{Mo}.

\subsection{Restricted Boltzmann Machine for Classification}
\begin{figure}[h]
    \centering
\includegraphics[scale=0.2]{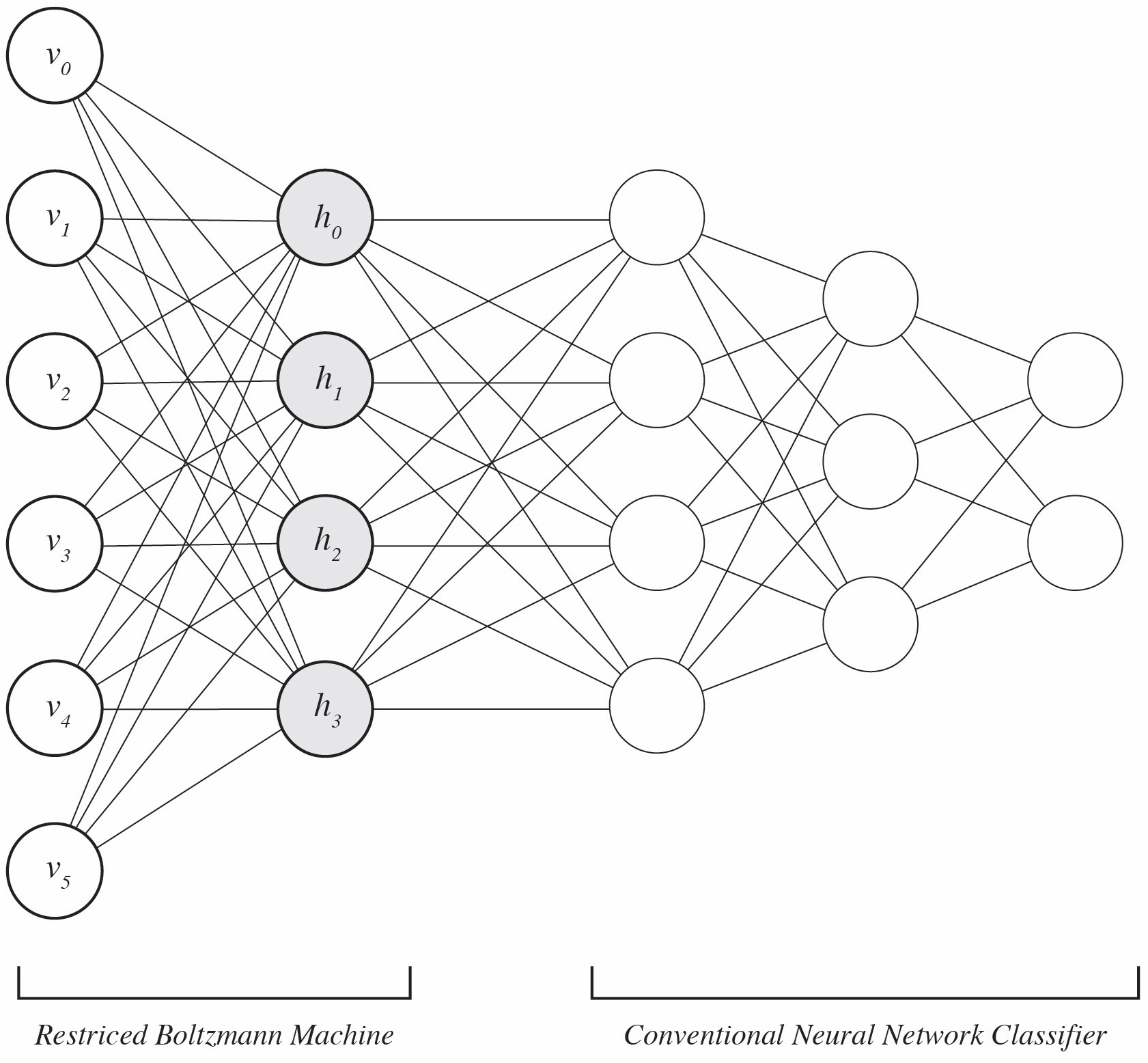}
\caption{Model of a Possible Architecture Used to Classify Samples Consisting of an RBM and Classifier}
\label{classifier} 
\end{figure}

Being able to reconstruct gene expression data is useful, yet it isn't immediately apparent how it can be used to classify different subtypes of lung cancer. As RBMs are not ordinary neural networks with well defined cost functions and conventional back-propagation they cannot be trained simply by employing usual methods. One method of classification is to use a neural network classifier that uses the hidden variables as the input. Please see figure \ref{classifier}. 

Another method, and the one used for our experiment is to clamp visible neurons with the class in which they belong and train the RBM to be able to reconstruct those clamped values\cite{classification}. For example, if a dataset contains 100 patients that are either male or female, and 50 variables: one would construct an RBM with 50 visible neurons + 2 visible neurons that will contain the class information (clamp), for a total of 52 visible neurons. When training, the 50 variables would be fed into the RBM as usual, with [1, 0] as the clamp value for male and [0, 1] for female. Thus the RBM, which is agnostic to the order of the data being fed in, would learn the distribution of data in relation to the clamp. In essence it would learn that when the 50 variables are distributed in a certain way the clamp is distributed in a unique way as well (particularly either [1, 0] or [0, 1]).

\begin{figure}[!h]
    \centering
\includegraphics[scale=0.14]{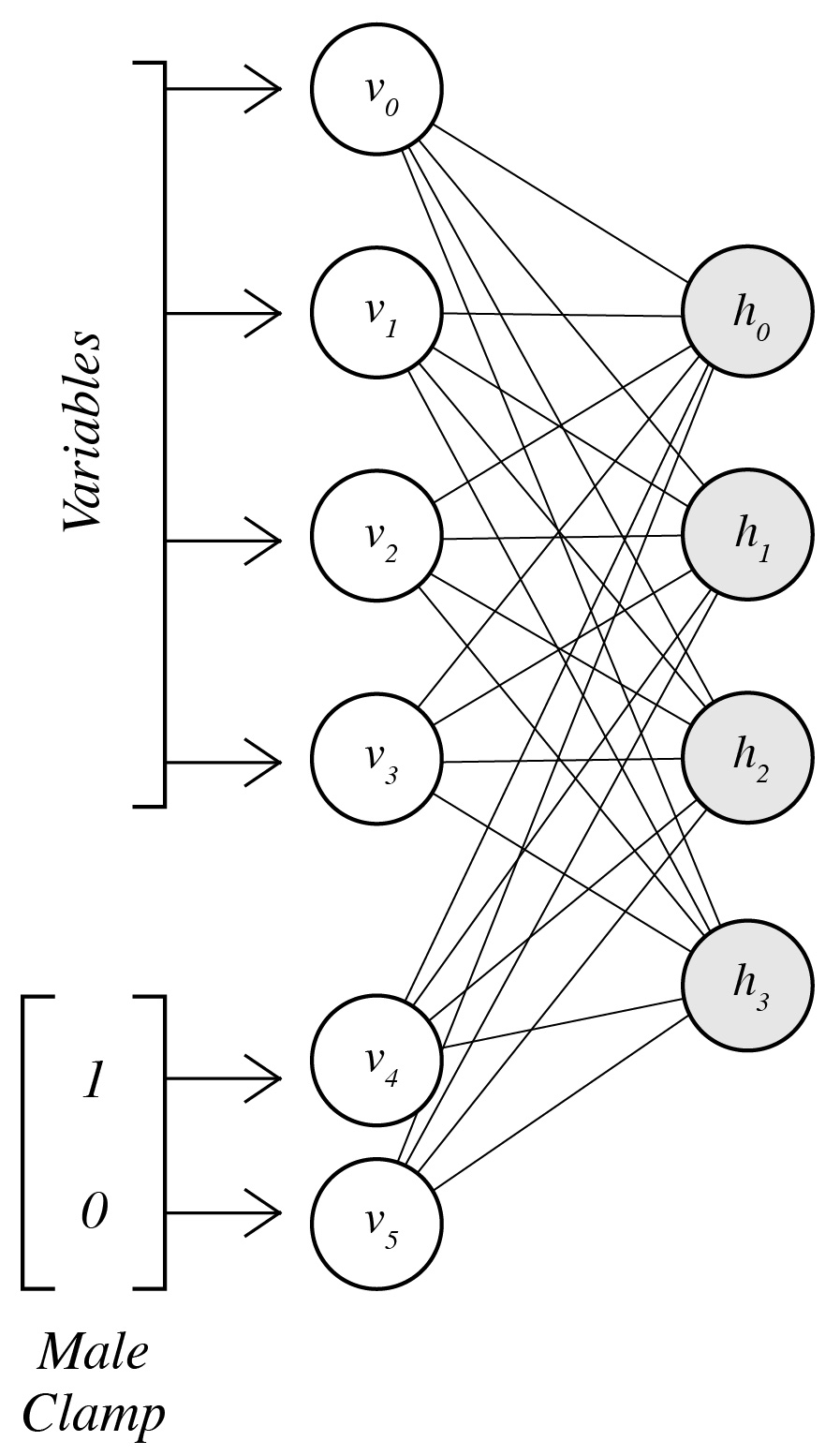}
\caption{Method of Classifying Samples Using a Clamp}
\label{clamp}
\end{figure}

Once the model is trained with a clamp (see figure \ref{clamp}), its accuracy can be validated by feeding a new patient vector with a neutral clamp [1, 1] and evaluating the result by collapsing the larger of the clamp values to a 1 and the other to a 0. E.g. if a 50 variable patient vector is fed into the RBM with a neutral clamp [1,1], the reconstruction would be a \textit{fuzzy} reconstruction of the patient vector with the clamp value [0.23, 0.48] which we would collapse to [0, 1], i.e., prediction would be female. It is important to note that you need an equal number of clamped neurons as the number of classes present in the dataset in order to provide a one-hot encoded clamp.

\section{Mapping A Boltzmann Machine To The QPU}
With a better understanding of an RBM we can begin to understand the basis of how we utilize the D-Wave machine to train the RBM. At the simplest level, the chimera structure allows one to map the classical RBM to the QPU. A difference now is that the Hamiltonian of this new system consists of an energy where the spin terms in $E(h,v)$ above are replaced by operators\cite{qbm}, however the same three parameters exist for this mapped system as the classical RBM we defined. In essence, to train a generative model you must have some insight into the underlying probability distribution of the data. Since determining the true distribution is a computationally expensive task we sample from the distribution instead. For example, if you want to know the underlying distribution of the sum of two dice throws you can sample from the distribution to gain further insight. Samples would be like: 2, 6, 7, 9, 7, 11, 3, 6, etc. Given enough samples you would be able to better construct the true probability distribution of the sum of two dice throws.

\begin{figure}[h]
    \centering
\includegraphics[scale=0.12]{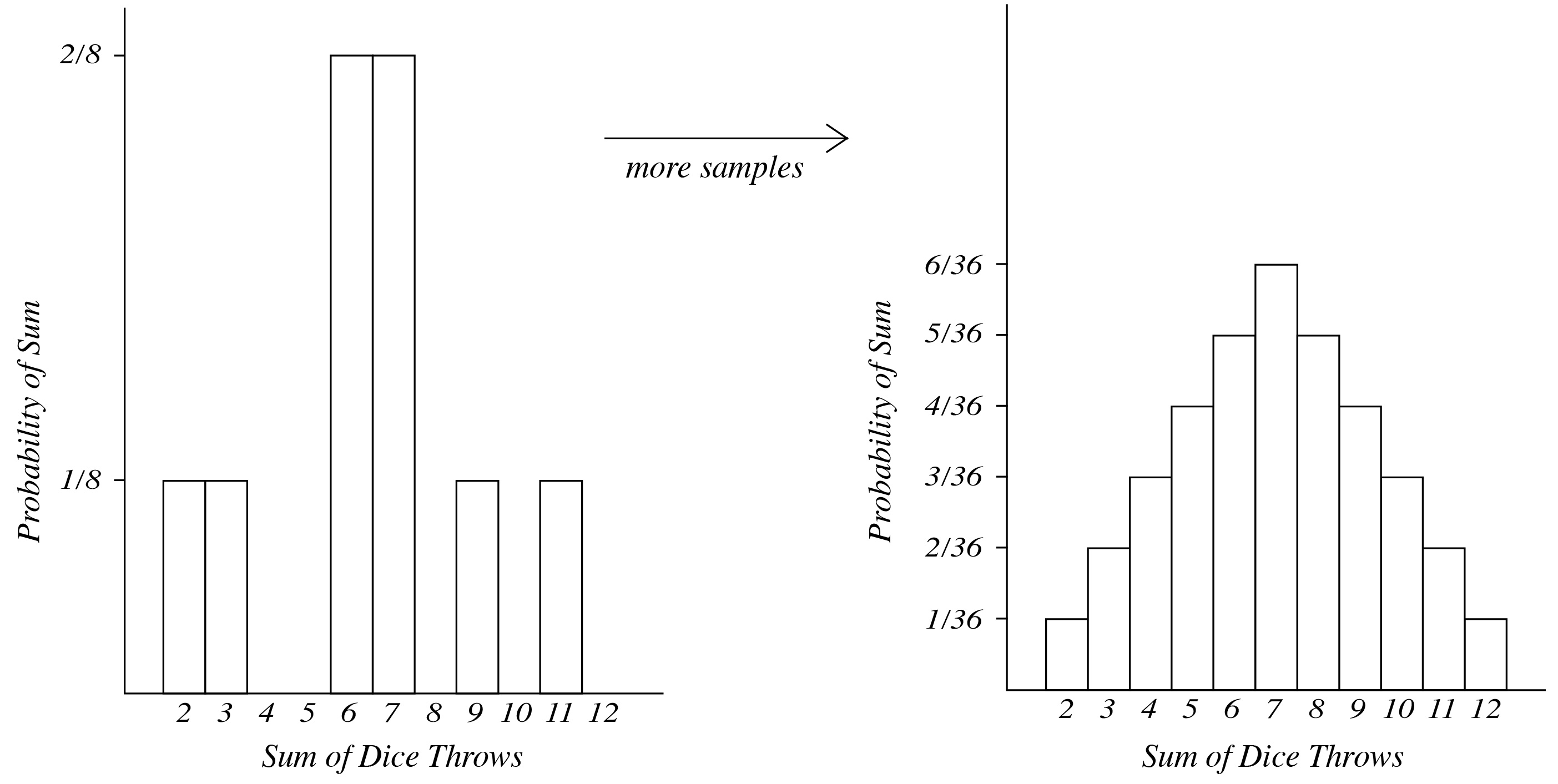}
\caption{Histogram of the Sums of Two Dice Being Thrown Trials}
\label{dices}
\end{figure}

This process of sampling scales to larger, more complex multivariate probability distributions. The QPU has an ability to sample from a more complex distribution and this is due to the fact that the quantum mechanics that drives the QPU inevitably introduces fluctuations that act as perturbations that can keep the system out of its ground state. This `restlessness' provides a complex landscape from which to sample and thus to enrich the training of the RBM. 

We will use the QPU to conduct the classification of lung cancer,  but first we must reduce the dataset from roughly 20,000 gene expressions down to a more manageable amount.

\subsection{Feature Selection}
We independently used three main methods of reducing data, namely univariate variable reduction, XGBoost, and, QCrush. The primary reason for reducing down the number of variables is due to the physical restraints of the QPU. In addition, we do so to reduce noise from the set and be able to train solely with the most significant variables/indicators. Feature selection can be a very complex and important part of the machine learning process and there are many methods that are generally considered either filter, wrapper, or embedded methods \cite{featureselection}, and then combinations of these which we refer to as ensemble feature selection methods \cite{featureselection2}. For this project we used filter, wrapper, and embedded methods, specifically we experimented with the Fisher score, XGBoost \cite{boost}, and LASSO. For this experiment this made little difference and the results we report used the Fisher score. This reduced the number of genes from approximately 20,000 to three different sets with 10,20, and 50 genes. The results reported below is derived from the set with 10 variables. 

An interesting approach that we developed internally was referred to above as QCrush. We will review this briefly as it will be the subject of a future publication after we perform more experiments with it. Essentially what QCrush does is that it compresses many variables into a representation for each patient. This enriches how much information can be used to represent an object that is to be modeled, which in this case is a patient. The reason we created such an algorithm is due to limitations in the architecture of the D-Wave QPU, which limits how many variables can be used to create a model. Similar approaches have been used including autoencoders \cite{autoencoders} to deal with this issue. QCrush has an advantage in that the encoding can be visualized in terms of how patients relate to each other. However, the point is that one can use a method like this to reduce the size of the data to ensure that an optimum amount of information is utilized for the learning phase of the RBM on the QPU.   

\subsection{Training a Quantum Boltzmann Machine}
Once the number of features has been reduced there are a set of steps that need to be taken to train the QBM. Steps 6 through 9 were repeated thrice for each set of hyper-parameters for greater precision.

\begin{enumerate}
\item Partition the dataset. In our case we will train on 80 samples, validate on 10, and test on 14
\item Normalize the dataset using a standard linear scaling to ensure every datapoint is between 0 and 1
\item Determine the number of hidden variables and number of samples to use
\item Initialize the quantum sampler
\item Binarize the dataset
\item Clamp the desired result
\item Pass the batches through for training
\item Evaluate results on the validation set
\item Score results on test set
\end{enumerate}
These steps lay out the basic procedure by which the Quantum Boltzmann Machine is trained. For our procedure we will focus on a the dataset created after performing an XGBoost.

\subsubsection {Partition the Dataset}
The dataset created via XGBoost contains 104 patients with 3 significant variables. In order to accurately validate the models’ performance the patients were partitioned into training, validation, and testing sets with a 80:10:14 split of the patients. 

\begin{figure}[h]
    \centering
\includegraphics[scale=0.11]{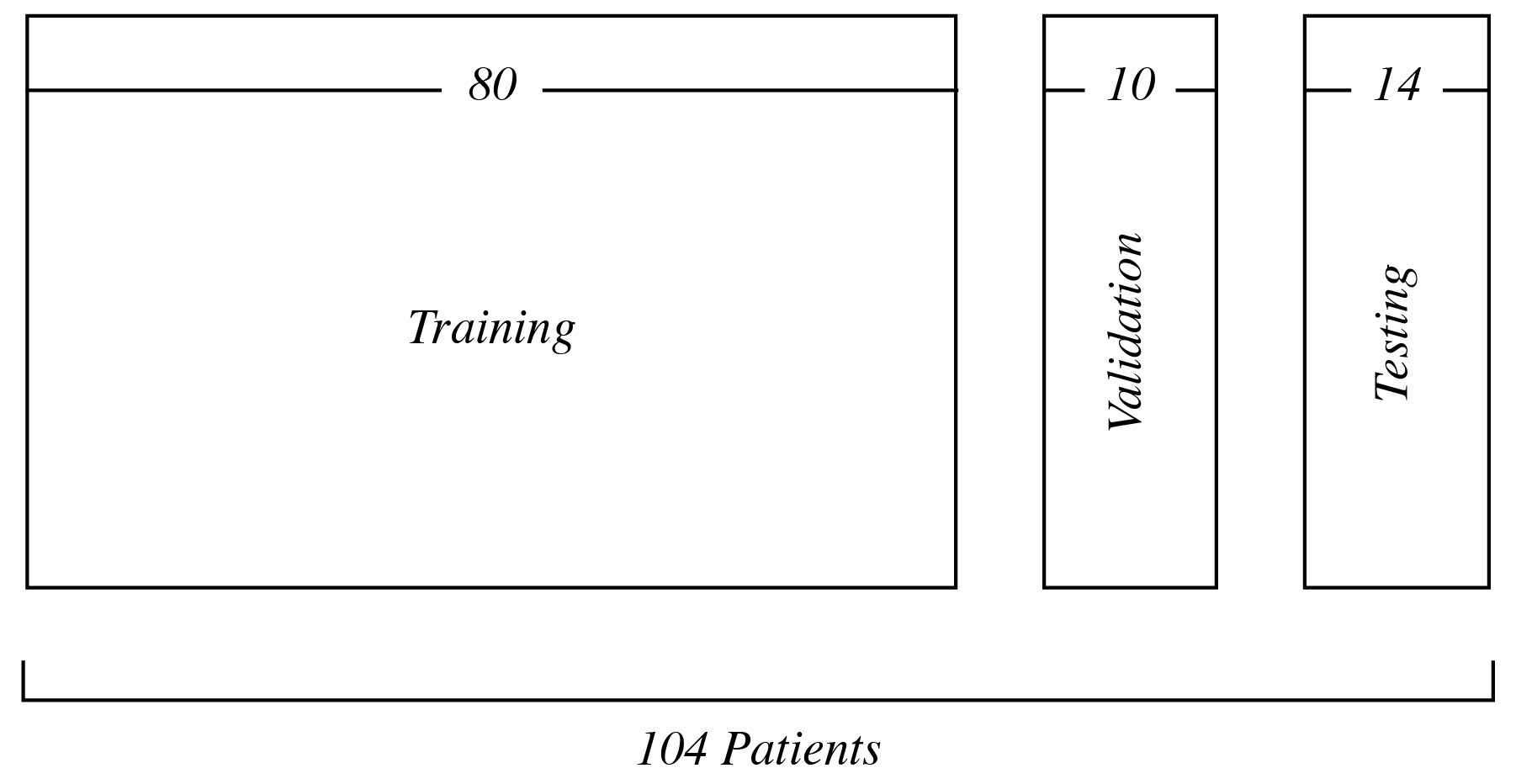}
\caption{Partitioning the Patient Dataset Into 80 Training, 10 Validation, and 14 Testing}
\label{partition}
\end{figure}

\subsubsection {Normalize the Dataset}
Restricted Boltzmann Machines are probabilistic models that seek to encode a complex probability distribution. By performing a linear normalization to the dataset we are able to utilize this probabilistic property of the model. Furthermore, when we binarize the dataset in order to represent it with binary samples, to be compatible with the QPU, it will be crucial that the values be contained within [0, 1]. This is because the algorithm that is utilized during this process by the QPU is known as quadratic unconstrained binary optimization (QUBO) \cite{qubo} and the vector quantities must be binary. The D-Wave QPU is designed to excel at solving these types of problems. 

We normalized in the following way: 

\[
\frac{\text {current element} - \text {minimum element}}{\text {maximum element} - \text {minimum element}}
\]

\subsubsection {Determine Hyperparameters}
The hyper-parameters for the QBM that we are expressly concerned with are:
\begin{enumerate}[i)]
  \item the number of samples
  \item the number of hidden nodes (i.e. data compression)
  \item the learning rate
\end{enumerate}With a rudimentary form of hyper-tuning, our approach was simply to iterate over every number of hidden layers from \numrange{1}{3}, learning rate from \numrange{0.25}{1.25} with step size 0.25, and samples from \numrange{1}{2048} in powers of 2. In the section dedicated to hyper-tuning we will speak to the results from our hyper-tuning.

\subsubsection {Initialize the Quantum Sampler}
Create an instance of the QBM class with the hyper-parameters that were determined. In addition the QBM is initialized with the number of visible nodes (number of features + classes)(refer to clamping for reasoning).

\subsubsection {Binarize the Dataset}
In order to utilize the Quantum Sampler the input data must contain only binary samples of the dataset, yet the data that is available is not so. Once normalized each patient is a vector of floating point data points between \numrange{0}{1}. To now binarize this vector of floats we broadcast a single vector into a set of one thousand vectors as follows. To illustrate with a broadcast to just ten vectors:

\begin{figure}[h]
    \centering
\includegraphics[scale=0.11]{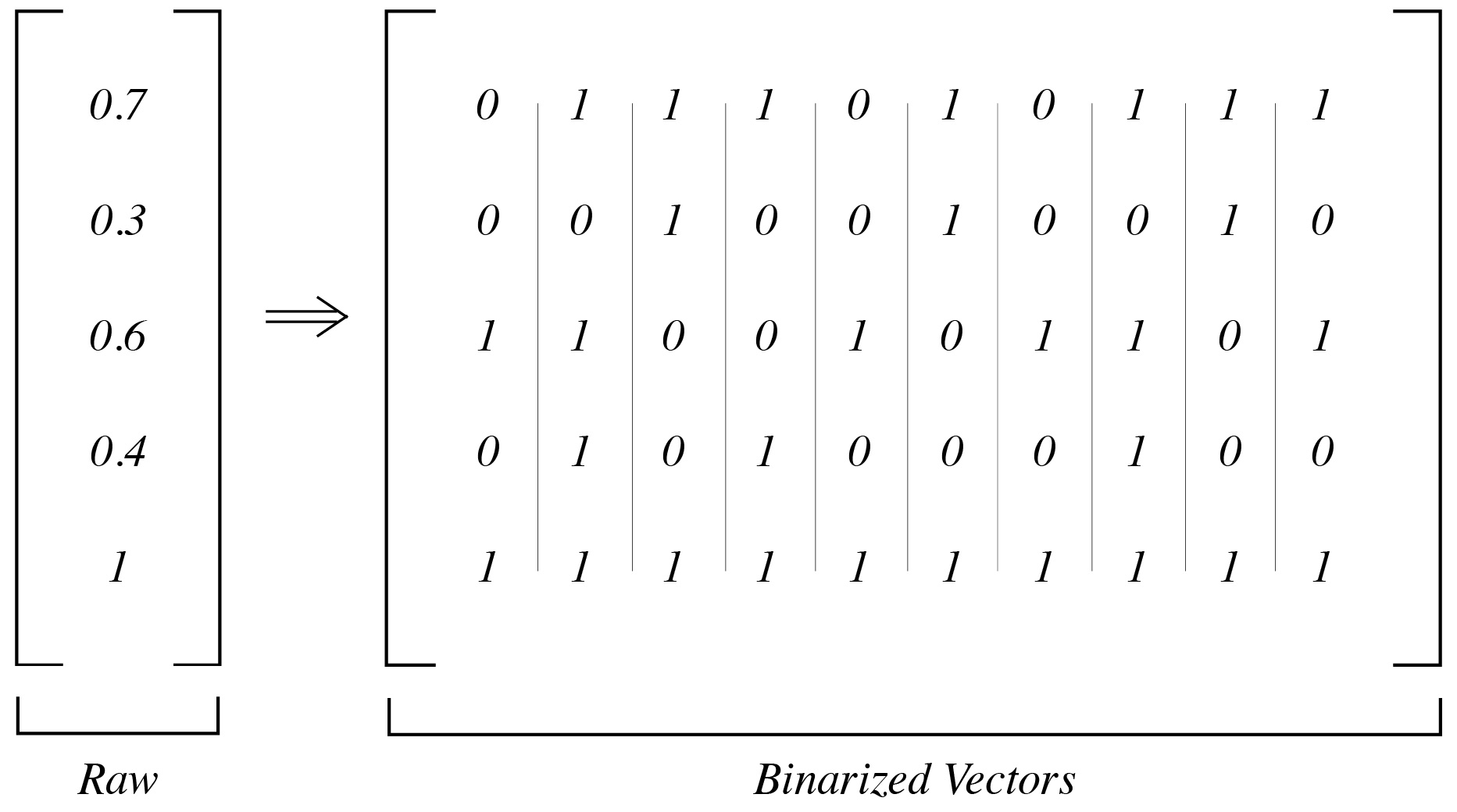}
\caption{Binarization of a Normalized Vector to a Set of Binary Vectors}
\end{figure}

Every value in a column vector is expanded from being a single value to a vector of one thousand 1s and 0s, where the number of 1s in the new vector correspond to the probability encoded by the single value. E.g. if the first variable of the patients' column vector is normalized to the value 0.7, the new binary vector representation of that single variable will be a thousand element vector with 700 ones, and 300 zeros in a random ordering. These 1000 column vectors are treated as a single batch when training the QBM, meaning each batch contains a single patient represented by 1000 binary vectors.

\subsubsection {Clamp the Class}
Adenocarcinoma and Squamous Cell Carcinoma are the 2 classes that the patient data belongs to. Thus we have a clamp of size 2 with [1,0] representing Adenocarcinoma and [0, 1] representing Squamous Cell Carcinoma. Once we append these clamp values to the patient vector we can begin to train.

\subsubsection {Training}
Every binarized batch of 1000 vectors is trained using the D-Wave quantum computer and the weights matrices are updated to be able to produce more accurate representations of the dataset. The operation of creating a batch of binarized test data and determining the hyper-parameters are in essence the only major requirements to train the QBM. Once the QBM is initialized and the dataset is processed into the aforementioned batches, D-Wave’s libraries handle the task of sampling from the QPU and using the gathered samples to train the model.

\subsubsection {Validation}
To validate the classification accuracy of the QBM we compute an error. This error is the euclidean distance between the vectors representing the clamp value i.e. [1, 0] and the predicted clamp value. In layman's terms, we determine an error by summing the square of the differences between each corresponding vector value of the predicted and true clamp. The lowest consistent error value indicates the best trained model with optimal hyper-parameters. To determine the predicted classification we repeat the same process for normalizing and preprocessing as before, yet instead of providing the clamp with the actual classification we provide a neutral clamp, i.e., [1, 1]. Then we simply feed the batch forward, and back, and calculate error.

\subsubsection {Testing}
Once validated we can test the QBM’s classification accuracy by repeating the same process as validation. The one difference between validation and testing is that we will not be testing for an error score, rather a raw value of success/failure in predication. If the models we create are over-fit we should observe a great disparity between the error found when validating and the actual test results.

\section {Results}
\begin{itemize}
\item the results of experiment showed that a learning rate of 0.75, 3 hidden neurons, and 1024 samples produced raw scores of 13, 14, and 13
\item the average score being \(\frac{40}{42} = 95.24\%\).
\item observed an increase in prediction accuracy with greater samples, as expected, and an increase in accuracy with a greater number of hidden nodes, with dimishing returns. 
\end{itemize}

\pgfplotsset{every axis/.append style={
                    label style={font=\tiny},
                    tick label style={font=\tiny} ,
                    }}
\begin{tikzpicture}

\begin{axis}[
    title={Histogram of Raw Score with Learning Rates},
    xlabel={Learning Rate},
    ylabel={Frequency of Raw Score on Test Dataset},
    xmin=0, xmax=14,
    ymin=0, ymax=30,
    xtick={0, 1, 2, 3, 4, 5, 6, 7, 8, 9, 10, 11, 12, 13, 14},
    ytick={0, 6, 12, 18, 24,30},
    legend pos=north west,
    ymajorgrids=true,
    grid style=dashed,
width=\textwidth*0.85,
]
 
\addplot[
    color=blue,
    mark=square,
   mark size=1.7pt
    ]
    coordinates {
    (0,0)(1,0)(2,0)(3,4)(4,3)(5,14)(6,17)(7,18)(8,15)(9,11)(10,1)(11,4)(12,4)(13,13)(14,4)
    };

 \addplot[
    color=red,
    mark=square,
    mark size=1.7pt
    ]
    coordinates {
    (0,0)(1,0)(2,1)(3,1)(4,2)(5,7)(6,6)(7,13)(8,13)(9,11)(10,6)(11,8)(12,13)(13,20)(14,7)
    };

\addplot[
    color=black,
    mark=square,
    mark size=1.7pt
    ]
    coordinates {
    (0,0)(1,0)(2,0)(3,1)(4,4)(5,3)(6,7)(7,8)(8,7)(9,9)(10,8)(11,7)(12,8)(13,29)(14,17)
    };

\addplot[
    color=orange,
    mark=square,
   mark size=1.7pt
    ]
    coordinates {
    (0,0)(1,0)(2,0)(3,2)(4,2)(5,8)(6,13)(7,16)(8,5)(9,7)(10,7)(11,6)(12,9)(13,17)(14,16)
    };

\addplot[
    color=green,
    mark=square,
   mark size=1.7pt
    ]
    coordinates {
    (0,0)(1,0)(2,1)(3,1)(4,4)(5,4)(6,9)(7,11)(8,9)(9,10)(10,5)(11,8)(12,14)(13,21)(14,11)
    };
    
\legend{0.25,0.50,0.75,1.00,1.25}

\end{axis}
\end{tikzpicture}

\section {Future Work}
Due to the limitations of time-sharing the D-Wave computer, in addition to constraints of the service itself, we were unable to test a greater number of hyper-parameters and conditions. In the future we hope to further explore the capabilities and potential optimizations of the D-Wave quantum system. That being said, we have already constructed interesting models to predict mild vs aggressive cases of Chronic Lymphocytic Leukemia (CLL). However, a hope that we have and see as a potential avenue to achieve true quantum supremacy in the near term involves the ability to explore the molecular and genetic landscape of patient populations. 

It is already understood and believed that quantum supremacy can be achieved on near term quantum devices in the area of drug design. This is due to the fact that molecules are quantum entities and quantum computers provide a computational space to perform simulations that is obviously more natural for such a venture. This effort is underway by the authors of this paper but we propose a related space where quantum supremacy may be possible: the aforementioned molecular and genetic landscape of patient populations. To elucidate consider that one can capture hundreds of miRNA data + ~ 20000 mRNAs + a large number of methylation data (hundreds of thousands) + single nucleotide polymorphisms (millions) + microbiome data. This is a monstrous amount of data about each individual and the challenge is to create algorithms that will help us understand which subset of these variables actually characterize a patient or person in a meaningful way. For example, is there a relatively small number of these variables, say approximately 10-30, that could define a specific sub-type of human that would respond particularly well to a new treatment of pancreatic cancer? This not only defines a subtype of human, but perhaps a specific manifestation of the disease. In this way, one would even be able to direct the activities of drug designers and thus a truly personalized approach to medicine may be possible. A hopeful perspective, but one tamed by the monstrous complexity involved with this kind of variable reduction. Approaches that are deemed black boxes cannot help with this task and thus one will need methods that hand over the subsets of variables, and effectively `explain' themselves. Quantum computation may be able to play a significant role here because of our ability to utilize quantum parallelism. Work in this direction that utilized the QUBO \cite{qubo} algorithm on a quantum annealer can be found in a white paper here \cite{quantumfeatureselection}. Much progress has to be made in quantum computational technologies before problems like this can be truly addressed however near term machines may be able to make real progress in this direction by allowing for a more complex and complete sampling space. This will allow novel algorithms a deeper reach into the space of these variable subsets. We are beginning to explore this direction through work with the D-Wave technology.

\section {Conclusion}

We utilized a quantum annealer, namely the D-Wave 2000Q, to create a model capable of predicting if non-small cell lung cancer patients either have adenocarcinoma or squamous cell carcinoma, two different varieties of this deadly disease. The ability to know this can have life altering treatment consequences, especially as cancer treatment matures and becomes more personalized. The variables used to train the model was gene expression data derived from tumor samples. Considering that the machine had access to 104 patients in total, the machine performed very well and we have reason to believe that the models are robust, as classical methods performed similarly and replicated well. The effort however was not made to demonstrate quantum supremacy of any kind, but to explore how precision medicine may be impacted as near term devices become more powerful. The ability to sample from complex distributions, something that quantum annealers are able to readily provide, is allowing generative based machine learning models, like ours, to become a reality. In \cite{neartermqc}, various approaches like this are explored but an important goal of this same paper was to suggest that near term quantum computers can be utilized to eventually out perform classical computers but only for certain types of problems and utilizing certain approaches. Our work here was performed in the same spirit and is an initiation point for an ongoing effort to explore how this new paradigm of machine learning will impact the medical space. Precision medicine is the hope of the future where treatment protocols will be tailored for specific subpopulations of patients. In order for this to become a reality, we will need to utilize machine learning protocols to understand the disease from the perspective of patient populations and the various manifestations these classically identified illnesses take. In addition to the new disease definitions that this effort will bring forth, a new understanding that will influence treatment paradigms will emerge. We believe that quantum machine learning, even near term devices, will have an impact.

\subsubsection {Author Acknowledgments }
The authors would like to thank D-Wave for allowing us access to their technology and the excellent training provided by their team. Thank you Dr. Peter Wittek, Colin J.E. Lupton,  and Dr. Abhi Rampal. We would also like to thank Creative Destruction Labs for having us participate in the 2017 CDL Quantum Machine Learning stream. Thank you to Songeun You for providing the figures. A thank you to Queen's University and Dr. Harriet Feillotter for advice and data sources. Finally, a special thank you to the whole NetraMark Corp team and all investors who helped make this endeavor a reality.

\bibliographystyle{unsrt}
\bibliography{mylib}

\end{document}